\documentclass[letterpaper, 10 pt, conference]{ieeeconf}  

\IEEEoverridecommandlockouts                              

\usepackage{booktabs} 
\usepackage[hidelinks]{hyperref}
\usepackage{multirow} 
\usepackage{subcaption}
\usepackage{algorithm} 
\usepackage{algpseudocode}
\usepackage{bm}
\usepackage{multirow} %
\usepackage{graphics} 
\usepackage{epsfig} 
\usepackage{amsmath, amssymb} 
\usepackage{xcolor}
\usepackage[nolist]{acronym} 
\usepackage{cite}
\newacro{dof}[DoF]{Degree of Freedom}
\newacro{rl}[RL]{Reinforcement Learning}
\newacro{sdqn}[SDQN]{Sequential DQN}
\newacro{srl}[SRL]{Sequential RL}
\newacro{qdp}[QDP]{Quasi-Dynamic Parameterisable manipulation}
\newacro{mvp}[MVP]{Max Value Map}

\algrenewcommand\algorithmicrequire{\textbf{Input:}}
\algrenewcommand\algorithmicensure{\textbf{Output:}}

\def\I{\mathcal{I}}
\def\L{\mathcal{L}}
\def\M{\mathcal{M}}
\def\R{\mathbb{R}}
\def\E{\mathbb{E}}

\def\N{\mathbb{N}}

\def\v{\mathbf{v}}

\def\a{\mathbf{a}}

\def\s{\mathbf{s}}
\def\v{v_\text{mid}}
\def\pick{\mathbf{a}_\text{pi}}
\def\place{\mathbf{a}_\text{pl}}
\def\param{\mathbf{a}_\theta}

\DeclareMathOperator*{\argmax}{arg\,max}

\newcommand{\eg} {\textit{e.g.,}~} %

\newlength{\Oldarrayrulewidth}
\newcommand{\Cline}[2]{%
	\noalign{\global\setlength{\Oldarrayrulewidth}{\arrayrulewidth}}%
	\noalign{\global\setlength{\arrayrulewidth}{#1}}\cline{#2}%
	\noalign{\global\setlength{\arrayrulewidth}{\Oldarrayrulewidth}}}

\graphicspath{{img/}}

\title{\LARGE \bf
QDP: Learning to Sequentially Optimise Quasi-Static and Dynamic Manipulation Primitives for Robotic Cloth Manipulation
}
\author{David Blanco-Mulero$^{1}$, Gokhan Alcan$^{1}$, Fares J. Abu-Dakka$^{2}$, Ville Kyrki$^{1}$
\thanks{This work was financially supported by the Academy of Finland grant numbers 317020 and 328399. Abu-Dakka, F. is supported by the European project euROBIN under grant agreement No 101070596.}
\thanks{$^{1}$David Blanco-Mulero, Gokhan Alcan and Ville Kyrki are with School of Electrical Engineering, Aalto University, Finland.) (e-mail: \texttt{david.blancomulero@aalto.fi}}%
\thanks{$^{2}$Munich Institute of Robotics and Machine Intelligence, Technische Universität München, 80992 München, Germany. Part of the research presented in this work has been conducted when Abu-Dakka, F. was at Intelligent Robotics Group, EEA, Aalto University, 02150 Espoo, Finland}
}

\begin{document}

\maketitle
\thispagestyle{empty}
\pagestyle{empty}

\begin{abstract}
Pre-defined manipulation primitives are widely used for cloth manipulation.
However, cloth properties such as its stiffness or density can highly impact the performance of these primitives.
Although existing solutions have tackled the parameterisation of pick and place locations, the effect of factors such as the velocity or trajectory of quasi-static and dynamic manipulation primitives has been neglected.
Choosing appropriate values for these parameters is crucial to cope with the range of materials present in house-hold cloth objects.
To address this challenge, we introduce the Quasi-Dynamic Parameterisable (QDP) method, which optimises parameters such as the motion velocity in addition to the pick and place positions of quasi-static and dynamic manipulation primitives.
In this work, we leverage the framework of Sequential Reinforcement Learning to decouple sequentially the parameters that compose the primitives.
To evaluate the effectiveness of the method we focus on the task of cloth unfolding with a robotic arm in  simulation and real-world experiments.
Our results in simulation show that by deciding the optimal parameters for the primitives the performance can improve by 20\% compared to sub-optimal ones.
Real-world results demonstrate the advantage of modifying the velocity and height of manipulation primitives for cloths with different mass, stiffness, shape and size.
Supplementary material, videos, and code, can be found at \texttt{\url{https://sites.google.com/view/qdp-srl}}.
\end{abstract}

\IEEEpeerreviewmaketitle

\section{Introduction}

The broad variety of fabric materials that are handled by humans everyday presents several challenges when manipulated by robotic systems.
As an example, the variability of cloth materials in house-hold objects in terms of shape, stiffness, elasticity, and mass~\cite{garcia_2022_cloth_set}, requires humans to perform a set of diverse manipulation actions such as those used for dressing assistance~\cite{joshi_2017_clothing_assistance_dmps} or flattening, folding and twisting cloths~\cite{biao_2019_imitation_cloth_randomforest}. 
However, it is not trivial to transfer such skills to robot manipulators, as manipulation primitives are often manually designed for a specific application~\cite{ha2021flingbot}, or a set of pre-tuned primitives is used~\cite{yahav_2022_speedfolding}.
Thus, to cope with a wide range of cloths, robotic systems should integrate learning algorithms that autonomously decide the parameter values of these manipulation primitives.

\begin{figure}
\vspace{0.2cm}
	\centering
	\def\svgwidth{\linewidth}
	{
 \fontsize{8}{8}
		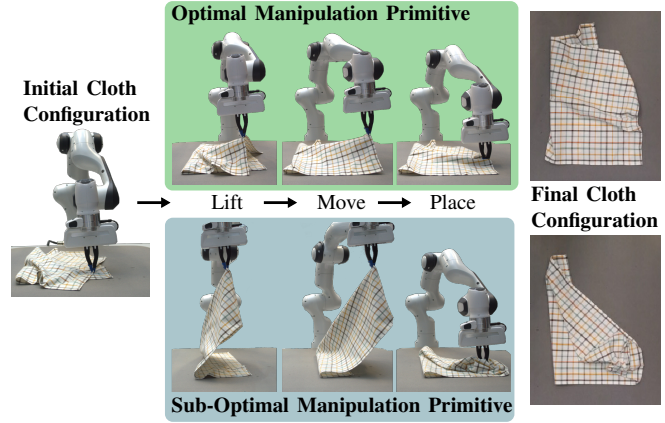}
	\caption{\acs{qdp} sequentially optimises the manipulation primitive parameter values to achieve better cloth configurations (\textit{green}) compared to using sub-optimal parameter values (\textit{blue}) for a manipulation primitive such as pick-and-place.}
	\label{fig:fancy_fig}
\end{figure}

The manipulation primitives used in robotic cloth manipulation fall into two categories: \textit{quasi-static}, such as the pick-and-place primitive~\cite{wu2019learning}; and \textit{dynamic}, which involve the forces of acceleration of the manipulator~\cite{Mason_2001_dynamic_quasistatic_boodk}, such as the fling primitive~\cite{ha2021flingbot}.
Quasi-static primitives have been extensively used for cloth folding and unfolding~\cite{yahav_2022_speedfolding, wu2019learning, lee_2020_one_hour_real, lin2021_vcd, Huang_2022_mesh_based_dynamics, canberk_2022_clothfunnels}, where different algorithms have been proposed for deciding the end-effector pick-and-place positions.
However, parameters such as the specific trajectory height or velocity of the primitive have been neglected.
This is crucial in applications such as cloth manipulation in-contact with a surface, where the size of the cloth will drastically affect the manipulation result, \eg bigger cloths will have more contact if they follow a trajectory with lower height (see Fig~\ref{fig:fancy_fig}).
Hence, these parameters should be taken into account to cope with a diverse set of cloth materials and sizes.

The main contribution of this work is a method to learn a visual policy that can determine the optimal parameters of manipulation primitives for cloth manipulation. 
We refer to our method as \acs{qdp}, short for the \acl{qdp} method.
Our key idea is to find the optimal parameters by following a sequential decision approach.
We take inspiration from the \ac{srl} framework \cite{wang_2021_sequential_rl_se3} to sequentially assign an action space to each primitive parameter.
Here,  each parameter of the primitive is informed by the previous proposed parameter.
This enables learning the relationship between the primitive parameters and their impact on the manipulation performance.
Our method works in a joint action space: the spatial action space of pick-and-place locations; and the parameter action space, that defines parameters such as the velocity of the manipulation primitive.
In order to evaluate the effectiveness of the sequential decision of parameter values, we evaluate our proposed method in simulation and real-world experiments in the task of cloth unfolding by using a single robotic arm.
Our contributions include:
\begin{itemize}
    \item Introducing \acs{qdp}, a novel approach that can optimise the parameters of manipulation primitives, decoupling the pick-and-place decisions as well as additional primitive parameters, without supervision or hand-labeled data during training,
    \item An analysis of the performance of \acs{qdp} on the cloth unfolding task, showing the superior performance of the proposed method compared to baselines, as well as its ability to find optimal parameters for both quasi-static and dynamic manipulation primitives,
    \item For the first time, a real-world evaluation of different manipulation primitives on a public cloth unfolding benchmark.
\end{itemize}

\section{Related Works}
\textbf{Cloth Manipulation} research has extensively focused on visual-based methods to plan the manipulation actions for a variety of tasks such as folding or unfolding~\cite{yahav_2022_speedfolding, wu2019learning, lee_2020_one_hour_real, lin2021_vcd, yamakawa2011dynamic}.
Prior work has actively studied how to decide the pick-and-place locations of pre-defined manipulation primitives.
Recent works have proposed to use spatial action maps~\cite{Wu_2020_SpatialActionMaps} to decide the pick position, and rotating and scaling the input image to decide the place position~\cite{ha2021flingbot, lee_2020_one_hour_real, canberk_2022_clothfunnels}. 
However, the decision of the place position in these approaches is limited to a discrete combination of rotation angles and scales, thus restricting  the primitive action space. 
In contrast, our work treats both pick and place as actions in a spatial action space, resulting in a more flexible primitive action space.
Alternatively, other works do not suffer from a restricted place location by defining a conditional action space~\cite{wu2019learning}, or action spaces that contain both the pick and place locations~\cite{lin2021_vcd, Huang_2022_mesh_based_dynamics, weng_2021_fabricflow}.
However, all these works suffer from the same caveat, they focus on determining only the pick and place of pre-defined manipulation primitives, neglecting other parameters such as the velocity at which they are executed.
The velocity at which the manipulation actions are executed has an impact in tasks such as cloth folding \cite{Hietala_2022_ours}, where dynamic manipulation actions can be adapted according to the fabric material.
In this work, we decide parameters such as the velocity for a dynamic manipulation primitive, and the height for a pick-and-place primitive, thus generating a more diverse set of manipulation actions to cope with cloths of varied sizes and materials.

\textbf{Sequential RL} was first introduced by Metz et. al~\cite{metz_2017_discrete_sequential_rl} to enable learning in high dimensional spaces by decomposing the action space in simpler sub-action spaces, proposing the \ac{sdqn} method.
This work was then extended to learn policies in SE(3) action spaces~\cite{wang_2021_sequential_rl_se3}, where they introduced the idea of augmented state that we follow in the proposed method.
The augmented state approach has shown success applied to learning grasping~\cite{Zhu_2022_equivariant_grasping} and manipulation~\cite{wang_2022_equivariant_q}.
Although prior works have extended the action spaces to SE(2) and SE(3), they have not considered other action spaces such as the manipulation primitive parameter action space. 
In this work, we follow the \acs{sdqn} method to compose an action space of pick and place location, as well as a parameter space that can modify the manipulation primitives behaviour; showing its effectiveness in cloth manipulation. In addition, prior work computed the pick and place actions at different time steps, whereas in this work we sequentially decide both, thus deciding the place location without interacting with the environment.

\begin{figure*}[t]
\vspace{0.2cm}
	\centering
	\def\svgwidth{1\linewidth}
	{
 \fontsize{7}{7}
		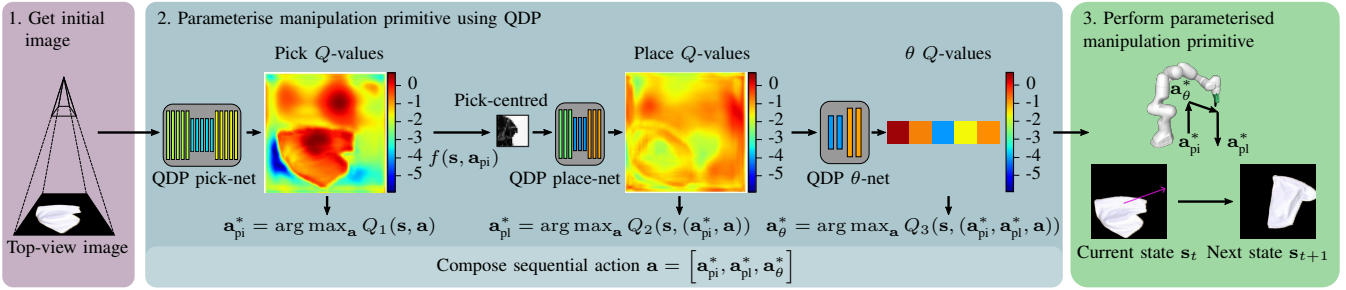}
	\caption{The proposed Quasi-Dynamic Parameterisable approach starts by getting a top-view image of the cloth. Then, to find the optimal manipulation primitive parameters, a sequential action $\a$ is composed from the sub-action output of three different networks: QDP pick-net, predicts the optimal pick position $\pick^*$; QDP place-net, predicts the place location $\place^*$; and QDP $\theta$-net, predicts additional primitive parameters $\param^*$, such as the primitive velocity. Each sub-action takes into account the previous information via encodings, e.g. the place sub-action accounts for the pick location using a pick-centred image. Finally, the manipulation primitive is executed with the optimal parameters placing the cloth on a new state.}
	\label{fig:proposed-method}
\end{figure*}

\section{Background}\label{sec:formulation}

\subsection{Augmented State Representation}\label{sec:formulation-sequential-rl}
The \acs{srl} augmented state approach introduced by Wang. et al.~\cite{wang_2021_sequential_rl_se3} is formulated as follows.
We define an MDP $\M = \langle S, A, R, T, \gamma \rangle$ with state space $S$, action space $A$, reward function $R$, transition dynamics $T$ and discount factor $\gamma \in [0, 1]$.
We denote this MDP as the \textit{top-level} MDP. The top-level MDP action space is composed by a set of \textit{sub-action} spaces $A=A_1 \times \cdots A_n$, where $n$ is the number of sequential actions and $\a_i \in A_i$ is an action in the $i$-th sub-action space.
Thus, the actions $\a = \left[ \a_1, \cdots, \a_n \right] \in A$ are composed by a set of multiple sequential choices.
A new MDP is defined by an augmented state representation $\bar{\M} = \langle \bar{S}, \bar{A}, R, T, \gamma \rangle$, with action space $\bar{A}=A_1 \cup \cdots \cup A_i$ and state space $\bar{S} = S_1 \cup \cdots \cup S_i$.
The decision making process starts with the initial \textit{sub-state} space $S_1 = S$ and sub-action space $A_1$.
The consecutive sub-state spaces take into account the sequence of previous sub-action spaces $S_i = S \times A_1 \times \cdots A_{i-1}$ for $i \geq 2$.
Following this formulation, the transition $p(\s'|\s, \a)$ can be rewritten as $p(\s' | \s, (\a_1, \cdots, \a_n))$.
Furthermore, the n-th transition has the same reward as the MDP $\M$, whereas the previous transitions where $i < n$ have zero rewards.
As an example, for a sequence of $n=3$ actions, the partial transitions $(\s, \a_1)$ and $(\s, \a_1, \a_2)$ have zero rewards, whereas the last transition $(\s, \a_1, \a_2, \a_3)$ has the same reward as the top-level MDP.

\subsection{Sequential DQN}\label{sec:formulation-sequential-dqn}
\acl{sdqn}~\cite{metz_2017_discrete_sequential_rl} decomposes the action space $A$ into $N$ sequential sub-actions to decrease computational complexity. 

The action-value functions are defined as $Q(\s, \a)$ for the top-level MDP, and as $Q_i(\s, \a_i)$ for the low-level MDPs.
In order to have consistent $Q$-values between the top-level and the low-level MDPs, \cite{metz_2017_discrete_sequential_rl} use a discount factor of $\gamma = 0$ for every $i < n$.
Thus, for a sequence of states in the top-level MDP $\s, \s' \in S$, the action-value functions satisfy
\begin{equation}
\begin{aligned}
    Q(\s, \a) &= Q_1(\s, \a_1),\\
    Q(\s', \a) &= Q_n(\s, \a_n) = Q_n(\s, (\a_1, \cdots, \a_n)).
\end{aligned}
\end{equation}
Here, the action-value functions are approximated by neural networks. The structure of the neural-networks proposed in this work is explained in detail in Section~\ref{sec:method}.

\section{Method}\label{sec:method}
\acs{qdp} learns a visual policy that can determine the optimal parameters of a manipulation primitive for cloth manipulation (see Figure~\ref{fig:proposed-method}).
Given an initial top-view image, our proposed approach determines the optimal pick, ${\pick^* \in A_1}$; place, ${\place^* \in A_2}$; and other parameters of the manipulation primitive, $\param^* \in A_3$.
These sub-actions are determined sequentially for computational feasibility. 

\subsection{Sequential Parameter Choice via SDQN}
The state is defined as a gray-scale image $\s \in \R^{D \times D}$, where $D$ is the height and width dimension of the image.
By decomposing the pick and place into multiple decisions we can decide the place position based on the pick action.
Thus, the first decision in our sequential RL setting is the pick position which is determined as 
\begin{equation}
    \pick^* = \argmax_{\pick} Q_1(\s, \pick),
\end{equation}
where the optimal pick $\pick^* \in \N^2$ is the pixel position in the image space.
The action-value function $Q_1$ is approximated by a neural network that also provides an encoding $g(\s)$ of the state-space.
Then, for deciding the place position we use the augmented state $S_2=S \times A_1$.
In order to encode the sub-action $\pick$ we use a mapping $f: (\s, \pick) \mapsto \I_\text{pick}$ that creates a pick-centred image $\I_\text{pick} \in \R^{E \times E}$, which is a cropped image centred in the pick position, similar to \cite{wang_2021_sequential_rl_se3}.
Instead of using the state $\s$ as input to the action-value function $Q_2$ we use an encoding $g(\s)$, which is part of the output from the neural network that approximates $Q_1$. 
The place sub-action is then computed as
\begin{equation}\label{eq:place-q2-enconding}
    \place^* = \argmax_{\place} Q_2(g(\s), f(\s, \pick), \place),
\end{equation}
where $\place^* \in \N^2$, same as the pick sub-action.

Similarly, the last parameter sub-action augmented state $S_3 = S \times A_1 \times A_2$ re-uses information from both previous sub-actions. 
Here, the action-value function $Q_2$ is taken as input to provide information of the place sub-action. 
In order to re-use and share previous information, the third augmented state is given by the encoding $g(\s)$ as well as an encoding of the pick-centred image.
More details about the encodings are given in Section~\ref{sec:method-qdc-net}. 
The parameter sub-action is then computed as 
\begin{equation}
    \param^* = \argmax_{\param} Q_3(g(\s), f(\s, \pick), Q_2(\s_2, \place), \param),
\end{equation}
where $\param^* \in \N^1$, and the parameter values are defined in a different range of values for each manipulation primitive as detailed in Section~\ref{sec:exp-set-up}. 

Finally, the top-level MDP action is composed by the three sub-actions $\a = \left[\pick^*, \place^*, \param^* \right]$ that have been sequentially computed re-using previous information throughout the augmented state space.

\subsection{QDP-Net}\label{sec:method-qdc-net}
We denote the first part of the network as QDP pick-net, which follows a U-net structure \cite{ronneberger_2015_u_net} similar to \cite{wang_2021_sequential_rl_se3} with skip connections so as to not lose information throughout the encoder-decoder structure.
The QDP pick-net performs the mapping $h_1: \s \rightarrow (g(\s), Q_1(\s, \pick))$, providing the encoding $g(\s)$ as well as the Q-values for the pick sub-action.
This is followed by the QDP place-net mapping
\begin{equation}
    h_2: (g(\s), f(\s, \pick)) \rightarrow (w(f(\s, \pick)), Q_2(\s, \place)),
\end{equation}
which combines the Q-pick encoding with the pick-centred image to output a second encoding $w(f(\s, \pick))$, as well as the $Q$-values for the place sub-action via a decoder structure.

Finally, the last part of the network QDP $\theta$-net follows the mapping 
\begin{equation}
    h_3: (g(\s), w(f(\s, \pick)), Q_2(\s_2, \place)) \rightarrow Q_3(\s, \param),
\end{equation}
which re-uses previous encodings information and, after a convolutional and fully connected network structure, outputs the $Q$-values for the third sub-action.
Thus, the proposed network used in \acs{qdp} is the combination of the three mappings 
\begin{equation*}
    h: \s \xrightarrow[]{h_1, h_2, h_3} (Q_1(\s, \pick), Q_2(\s, \place), Q_3(\s, \param)),
\end{equation*}
which outputs the $Q$-values for the three sub-actions and sub-state spaces, used to determine the optimal sub-actions.

\subsection{Training procedure}\label{sec:method-training}
In this work we focus on the task of cloth unfolding. The objective is to maximise the area of the cloth to achieve a smooth or flattened state. Thus we define the reward for training the policy as
\begin{equation}
    r(\s) = C(\s)/C_\text{max},
\end{equation}
where $C(\s)$ is a function that computes the current area of the cloth based on the image pixels, and $C_\text{max}$ is the maximum area of the cloth.
Each of the $Q$-networks is optimised by minimising the Huber loss \cite{Huber1992}
\begin{equation}
    \L = \E (y - Q(\s, \a; \phi)),
\end{equation}
where $y$ is the optimisation target, $\phi$ denotes the parameters of the network used for evaluation, and $\psi$ the parameters of the \acs{sdqn} target network. The target is defined as the 1-step TD return
\begin{equation}
    y = r(\s) + \gamma Q_3(\s', (\pick', \place', \param'); \psi).
\end{equation}
Note that for all of the networks we use the same target, that is, the $Q$-values of the last $Q$-network, as it has been proven to be more sample efficient than using different targets per each network \cite{wang_2021_sequential_rl_se3}.

During training, in order to trade-off between exploration and exploitation we use $\varepsilon$-greedy.
In addition, to improve the sim-to-real transfer we also perform Domain Randomisation \cite{tobin_2017_domain_rand} and randomise the internal parameters of the cloth, including the stiffness, mass, size, colour, and initial cloth configuration.
Furthermore, the network parameters are initialised randomly and no prior is required.
Finally, in order to reduce the number of network parameters and based on the results from prior work on cloth manipulation \cite{Hietala_2022_ours}, we use a grayscale image $\s \in \R^{128 \times 128}$ as input to the network.
The observations are rotated at each time-step to help generalise the network to different cloth states. However, we do not augment the data by increasing or decreasing the image scale, as this would lose the spatial information that has been used for training the manipulation primitives.
Additional information is given in Section~\ref{sec:exp-set-up}.

\begin{figure}[t]
\vspace{0.2cm}
	\centering
	\def\svgwidth{0.9\linewidth}
	{
 \fontsize{12}{12}
\begingroup%
  \makeatletter%
  \providecommand\color[2][]{%
    \errmessage{(Inkscape) Color is used for the text in Inkscape, but the package 'color.sty' is not loaded}%
    \renewcommand\color[2][]{}%
  }%
  \providecommand\transparent[1]{%
    \errmessage{(Inkscape) Transparency is used (non-zero) for the text in Inkscape, but the package 'transparent.sty' is not loaded}%
    \renewcommand\transparent[1]{}%
  }%
  \providecommand\rotatebox[2]{#2}%
  \newcommand*\fsize{\dimexpr\f@size pt\relax}%
  \newcommand*\lineheight[1]{\fontsize{\fsize}{#1\fsize}\selectfont}%
  \ifx\svgwidth\undefined%
    \setlength{\unitlength}{762.99996924bp}%
    \ifx\svgscale\undefined%
      \relax%
    \else%
      \setlength{\unitlength}{\unitlength * \real{\svgscale}}%
    \fi%
  \else%
    \setlength{\unitlength}{\svgwidth}%
  \fi%
  \global\let\svgwidth\undefined%
  \global\let\svgscale\undefined%
  \makeatother%
  \begin{picture}(1,0.75884668)%
    \lineheight{1}%
    \setlength\tabcolsep{0pt}%
    \put(0,0){\includegraphics[width=\unitlength,page=1]{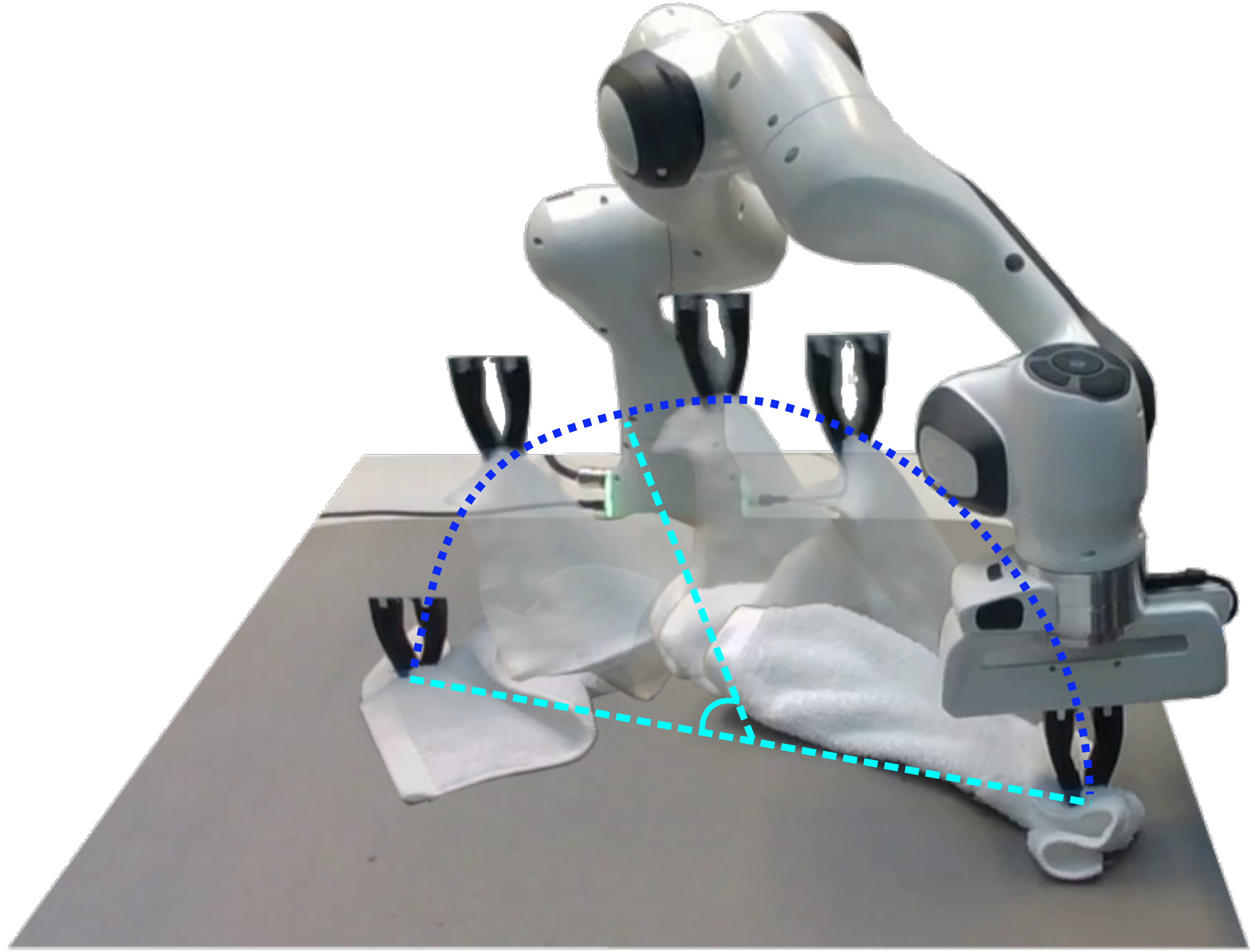}}%
    \put(0.52566754,0.39458545){\color[rgb]{0.04705882,0.16470588,0.98039216}\rotatebox{-10.116637}{\makebox(0,0)[lt]{\lineheight{1.25}\smash{\begin{tabular}[t]{l}$\bm{v}_\text{mid}^\text{des}$\end{tabular}}}}}%
    \put(0,0){\includegraphics[width=\unitlength,page=2]{QP_Fig_Design_v2.pdf}}%
    \put(0.23249036,0.22171938){\color[rgb]{1,1,1}\rotatebox{-0.9952857}{\makebox(0,0)[lt]{\lineheight{1.25}\smash{\begin{tabular}[t]{l}$\pick$\end{tabular}}}}}%
    \put(0,0){\includegraphics[width=\unitlength,page=3]{QP_Fig_Design_v2.pdf}}%
    \put(0.47131005,0.20598274){\color[rgb]{1,1,1}\rotatebox{-10.116637}{\makebox(0,0)[lt]{\lineheight{1.25}\smash{\begin{tabular}[t]{l}$\bm{\alpha(t)}$\end{tabular}}}}}%
    \put(0,0){\includegraphics[width=\unitlength,page=4]{QP_Fig_Design_v2.pdf}}%
    \put(0.87782089,0.13699576){\color[rgb]{1,1,1}\rotatebox{-0.9952857}{\makebox(0,0)[lt]{\lineheight{1.25}\smash{\begin{tabular}[t]{l}$\place$\end{tabular}}}}}%
  \end{picture}%
\endgroup%
}
	\caption{Dynamic manipulation primitive that moves from the proposed $\pick$ to $\place$ locations following a semi-circle shaped trajectory obtained from the quintic polynomial of $\alpha(t)$.
    }
	\label{fig:quintic-primitive}
\end{figure}

\section{Experiments}\label{sec:experiments}
Our experiments focus on two main points: 1) evaluating the performance of QDP compared to state-of-the-art baselines, and 2) evaluating different manipulation primitives on the task of cloth unfolding using a single robotic arm.
Thus, we analyse the following:
\begin{itemize}
    \item What is the impact of learning to sequentially optimise parameters such as the height or velocity of pre-defined manipulation primitives?
    \item How does the proposed network architecture for \acs{qdp} perform compared to other architectures?
    \item What is the performance of \acs{qdp} when transferred to the real-world in a zero-shot manner?
\end{itemize}

\subsection{Manipulation primitives}\label{sec:exp-manip-primitives}
We evaluate three manipulation primitives: pick-and-place, drag, and dynamic quintic polynomial; each defined at least by their pick and place locations.
The primitives are defined as: 
\begin{itemize}
    \item \textbf{Dynamic Quintic Polynomial}: dynamic manipulation primitive which is defined by its velocity. This primitive follows a semi-circle shaped trajectory from $\pick$ to $\place$, where $\v^{des} \in \left[0.08, 0.5\right]$ m/s is the desired velocity of the manipulator's end-effector at the middle point of the trajectory (see Fig.~\ref{fig:quintic-primitive}). The primitive is designed following a fifth-order polynomial of $\alpha$ angle \cite{spong2020robot}, expressed as
    \begin{equation}\label{eq:quintic-poly}
        \alpha(t) = a_0 + a_1t + a_2t^2+a_3t^3+a_4t^4+a_5t^5.
    \end{equation}
    \item \textbf{Pick-and-Place (P-n-P)}: quasi-static manipulation primitive with parameters $\{h_\theta, t_\theta\}$, 
    where the height has the range of values $h_\theta \in \left[0.1, 0.2, 0.3, 0.4, 0.5\right]$ m. to lift the cloth; and the time to move from $\pick$ to $\place$ is within the range of $t_\theta \in \left[10, 11, 12, 13, 14, 15\right]$ seconds. 
    \item \textbf{Drag}: quasi-static primitive with $t_\theta$ as additional parameter. This primitive can be seen as a P-n-P primitive where $h_\theta$ is equal to zero, and the time has the range of values as the P-n-P primitive.
\end{itemize}

\subsection{Baselines}\label{sec:exp-baselines}
In order to evaluate the performance of our method we compare against \ac{mvp}, proposed by \cite{ha2021flingbot}, which suggests the pick position based on a spatial action map \cite{Wu_2020_SpatialActionMaps}. The network takes a combination of 5 different scales and 12 rotations of the input image, selecting the combination with highest $Q$-value to define the place position.
It is important to note that the \ac{mvp} method iterates through the spatial action maps until it finds a valid pick position.
We use the same training procedure, where the input image is RGB $\s \in \R^{3 \times 64 \times 64}$, as in the original code provided by the authors. 
During training and evaluation of MVP, the primitives height, time, and velocity, are set by using the median of the values $\param$ proposed by QDP after training.

In addition, we evaluate the performance of the sequential decision of pick-and-place parameters using different network architectures. We compare the U-net architecture against two manipulation architectures. The first one is the network used in the \ac{mvp} method. The second network architecture is Form2Fit (F2F) \cite{zakka_2020_form2fit}, which showed great performance for rigid-body pick-and-place tasks.
It is important to note that these networks do not share any parameters to decide the action parameters.
Additional details about the implementation can be found on our website\footnote{\texttt{https://sites.google.com/view/qdp-srl}}.

\subsection{Experimental Set-Up}\label{sec:exp-set-up}

We employ Softgym \cite{lin2020softgym} as the simulation environment with the modifications done by \cite{ha2021flingbot}.
The simulation top-view images are originally in the dimension $\s \in \R^{3 \times 400 \times 400}$ and then reshaped onto the specific network input dimension.
We train and evaluate on the data sets provided by \cite{ha2021flingbot} to benchmark against an existing common baseline.

The training for all the baselines is done using a V100 GPU, which takes approximately 3 days.
All the methods are trained using three different random seeds until a total of $40\cdot 10^3$ environment steps are collected into the replay buffer.
We use the Adam optimiser \cite{Diederik_Adam_2015} and scale the rewards by a factor of 10. By increasing the magnitude of the rewards their variation is more pronounced facilitating the convergence of the networks.
The exploration coefficient starts at a value of $\varepsilon=1.0$ and decays $0.001$ per episode to finish at a value of $\varepsilon=0.01$ after 20000 episode steps.
During evaluation we rotate the top-view images in a range of $\{0, 180\}$ degrees, spaced by 15 degrees, and compare the pick $Q$-values of each input image.
This enables to select the action with the highest expected improvement under the same cloth configuration.

Our experiments evaluate the maximum coverage and coverage improvement that can be achieved within 10 interactions with the cloth, both normalised by the estimated maximum area of the cloth.
The coverage improvement allows to evaluate the performance of the methods when there is a high variance of the cloth initial configuration.

During evaluation, we use the depth image as prior data for correcting proposed pick points that are on the cloth low resolution image but are not in the high resolution image. Thus, pick points that are on a radius of 10 pixels of valid cloth points are used as pick location.
The real-world experiments are performed using an Azure Kinect DK to obtain both the top-view and depth images. Here, the top-view images are $\s \in \R^{3 \times 1280 \times 720}$, and reshaped accordingly.

For the real-world experiments we evaluate three different cloths from the public household cloth benchmark \cite{garcia_2022_cloth_set}.
We use a Franka Emika Panda to manipulate the cloths. Due to the limitations of the robot workspace we evaluate the objects that could be completely unfolded, which are the small towel ($0.3 \times 0.5$ m.), cotton napkin ($0.5 \times 0.5$ m.), and  chequered rag ($0.5 \times 0.7$ m.).
The initial cloth configuration is generated following the steps detailed by \cite{garcia_2020_benchmarking_cloth}, where the cloth is grasped from one of the corners and placed in a crumpled configuration on a foam mat.
Each of these cloths has a different pattern, size, elasticity, and weight; thus representing a wide variation of cloth types to evaluate the performance of each manipulation primitive.
The real-world manipulation primitives are performed using a Cartesian impedance controller, where the stiffness and impedance parameters have been tuned empirically for the Franka Emika Panda robot.

\begin{figure}
\vspace{0.2cm}
	\centering
	\def\svgwidth{\linewidth}
	{\fontsize{8}{8}
		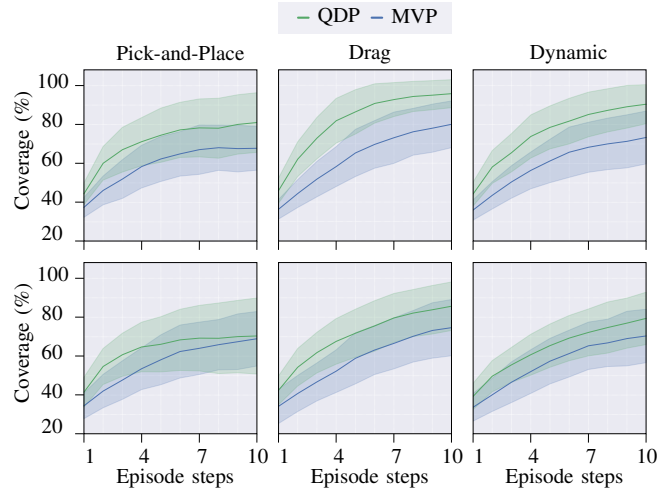}
	\caption{Quantitative comparison of the coverage percentage for unfolding cloths  in simulation. The results compare the proposed method QDP (\textit{green}) against the baseline MVP (\textit{blue}) for deciding the primitives parameter values. Three primitives are evaluated: Pick-and-Place, Drag, and Dynamic; where QDP selects the time $t_\theta$ and the velocity $\v$ of the quasti-static and dynamic primitives.
    The results show an increase in coverage using QDP for both normal size (\emph{Top}-row), and large size (\emph{Bottom}-row) cloths.
    }
	\label{fig:exp-sim-primitives}
\end{figure}

\section{Simulation Experiments}
\subsection{Manipulation Primitives}

\begin{figure*}
\vspace{0.2cm}
	\centering
	\def\svgwidth{\linewidth}
	{
 \fontsize{8}{8}
		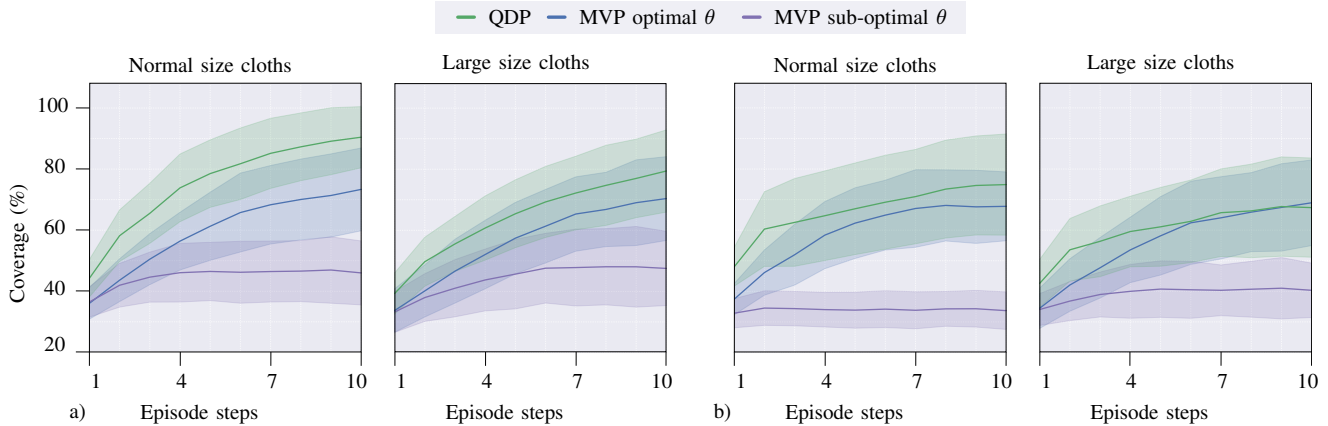}
	\caption{Quantitative comparison of coverage percentage for unfolding cloth in simulation using the a) Pick-and-Place, and b) Dynamic manipulation primitive.
    The results compare the proposed QDP (\textit{green}) against MVP using different velocity and height values. The additional parameter values are set as the median of the proposed values by QDP, denoted as MVP optimal $\theta$ (\textit{blue}), and a sub-optimal value (\textit{purple}); for normal rectangular cloths, and large rectangular cloths.
    This demonstrates that using sub-optimal primitive values is detrimental for the task of cloth unfolding.
    }
	\label{fig:exp-sim-sub-optimal}
\end{figure*}

We start by evaluating the performance of QDP against the MVP baseline for the aforementioned manipulation primitives. The coverage performance for both regular (\textit{top}) and large (\textit{bottom}) cloths test data sets is shown in Figure~\ref{fig:exp-sim-primitives}.
Here, MVP uses, as fixed parameters, $t_\theta=10$ for both P-n-P and drag, $h_\theta=0.2$ for P-n-P, and $\v=0.1$ for the dynamic primitive.
For all the primitives, QDP presents higher mean coverage, where the variance for both drag and dynamic primitives falls within the 100\% coverage.
Although the improvement on the large data set is lower, the increase in performance of QDP compared to the MVP baseline is consistent.
We attribute the reduced performance on the large data set to a low generalisation on the proposed method, as no large cloths are seen during training and the scale of the image is fixed as opposed to the baseline.

\subsection{Optimal and Sub-Optimal Primitive Parameter Values}
We analyse as well the effect of using optimal and sub-optimal velocities and height values for the manipulation primitives.
The results, in Figure~\ref{fig:exp-sim-sub-optimal}, show the performance of QDP against MVP using as fixed parameter the median of the proposed values by QDP; as well as an experimentally selected sub-optimal value.
In Figure~\ref{fig:exp-sim-sub-optimal} a), the dynamic manipulation primitive velocity has been set to an optimal value of $\v=0.1$, and a sub-optimal value of $\v=0.2$.
The results show that using a sub-optimal value for the dynamic manipulation primitive is detrimental, as the performance drops more than 20\% for both normal and large cloths.
We attribute this poor performance to constantly using a high velocity, as opposed to adapting it, for cloth sizes and configurations that do not require such acceleration to improve their coverage.

The results in Figure~\ref{fig:exp-sim-sub-optimal} b) compare the P-n-P primitive using an optimal height value of $h_\theta=0.2$, and a sub-optimal one of $h_\theta=0.5$, keeping the primitive time fixed to $t_\theta=10$. The constant large height results in a considerable performance drop for both normal and large size cloths. This is a consequence of lifting the cloth from a single point and losing contact with the surface, which results in a crumpled configuration when dropped onto its place location.

By using QDP to adapt the primitives the coverage performance improves at least 10\% for most of the cases.
We note that the P-n-P performance for large size cloths slightly drops compared to the optimal height, which can be a result of no large cloths in the training data set as mentioned before.

\subsection{Network Architecture}

\begin{figure}
	\centering
	\def\svgwidth{\linewidth}
	{
 \fontsize{8}{8}
		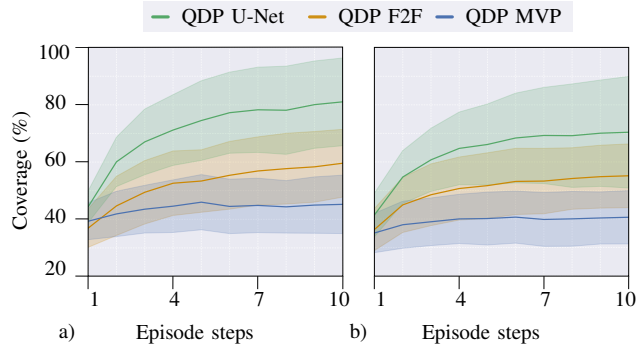}
	\caption{Quantitative comparison of coverage percentage for unfolding cloth in simulation using the Pick-and-Place manipulation primitive. The results compare three different architectures, the proposed method QDP U-Net (\textit{green}), Form2Fit or F2F (\textit{orange}), and MVP (\textit{blue}); for a) normal rectangular cloths, and b) large rectangular cloths.
 }
	\label{fig:exp-sim-nets}
\end{figure}

Next, we continue by analysing whether the proposed network structure is the right choice for the sequential decision process.
We evaluate the P-n-P primitive using different network architectures, shown in Figure~\ref{fig:exp-sim-nets}.
The U-Net architecture clearly outperforms the other two, showing an improvement of over 20\% for the F2F network, and up to 40\% for the MVP network.
This demonstrates that the proposed U-Net architecture, sharing information over the sequential decision process, is one of the key ingredients for the performance of \acs{qdp}.

\begin{figure*}
\vspace{0.2cm}
	\centering
	\def\svgwidth{\linewidth}
	{
 \fontsize{8}{8}
		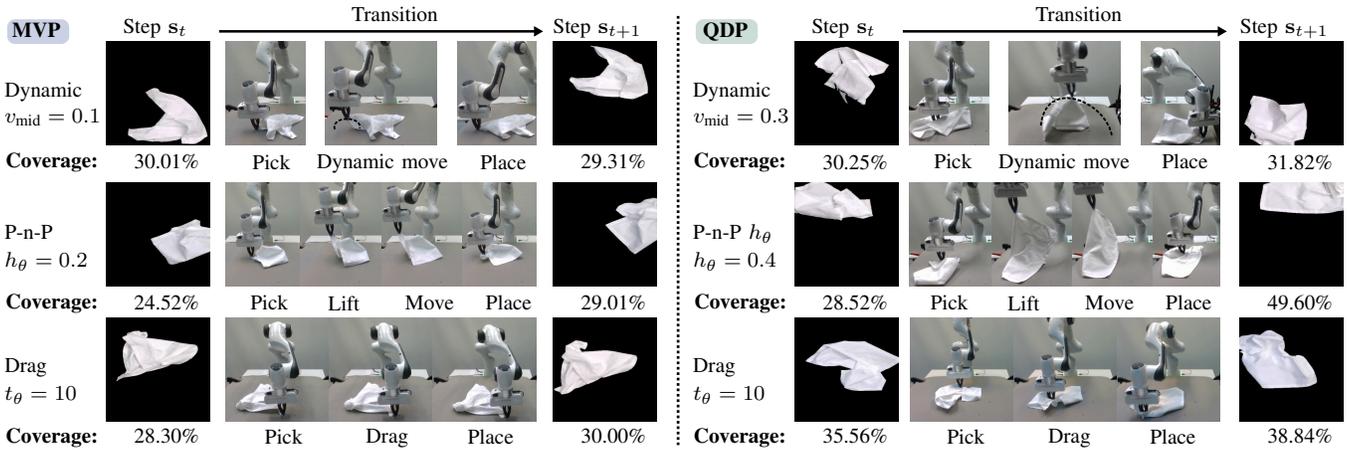}
	\caption{Qualitative results for the real-world experiments for unfolding the napkin from the house-hold cloth data set. The results show the coverage at each time step $\s_t$ and $\s_{t+1}$, as well as the trajectories using the dynamic, pick-and-place (P-n-P), and drag manipulation primitives for both the baseline MVP (\emph{left}) and the proposed method QDP (\emph{right}).
    The results show that by modifying the velocity and height parameters QDP can alter the cloth configuration and lead to a better coverage.
 }
	\label{fig:exp-real-world}
\end{figure*}

\section{Real-World Experiments}

\begin{table}
	\caption{Quantitative results of coverage improvement percentage after 10 interactions with the cloth in the real-world.
		The results show the performance of QDP and MVP for the the pick-and-place (P-n-P), drag and dynamic manipulation primitives. P-n-P $t_\theta$ and  $h_\theta$ refer to QDP proposing either the time or height of the P-n-P primitive, respectively.
  } 
	\resizebox{\linewidth}{!}{%
		\renewcommand\arraystretch{1.4} 
			\centering
			\begin{tabular}{c c | c|c|c|c}
				\Cline{1pt}{3-5}
				\multicolumn{2}{c}{ } & \multicolumn{3}{c}{Cloth} \\  \cline{3-5}
				\multicolumn{2}{c}{ }  &\multicolumn{1}{c|}{Towel} & \multicolumn{1}{c|}{Napkin} & \multicolumn{1}{c}{Chequered Rag} \\
				\noalign{\hrule height 1pt}
				\parbox[t]{2mm}{\multirow{3}{*}{\rotatebox[origin=c]{90}{{\textbf{MVP}}}}}&
				\multicolumn{1}{|c|}{P-n-P} & \multicolumn{1}{c|}{24.93 $\pm$	8.64} & \multicolumn{1}{c|}{17.69 $\pm$ 11.57}  &  \multicolumn{1}{c}{-0.30 $\pm$ 4.70} \\ \cline{2-5}
				&
				\multicolumn{1}{|c|}{Drag} & \multicolumn{1}{c|}{-1.21 $\pm$ 6.34} 
                    & \multicolumn{1}{c|}{3.72 $\pm$ 4.02} & \multicolumn{1}{c}{-6.41 $\pm$ 2.96}  \\ \cline{2-5}
				& 
				\multicolumn{1}{|c|}{Dynamic}
				&  
				\multicolumn{1}{c|}{9.89 $\pm$ 1.16} & \multicolumn{1}{c|}{1.57 $\pm$ 1.01} 
                    & \multicolumn{1}{c}{-6.04 $\pm$ 8.45} \\ 
					\Cline{1pt}{1-5}
				\parbox[t]{4mm}{\multirow{4}{*}{\rotatebox[origin=c]{90}{\parbox{.7cm}{\textbf{QDP (Ours)}}}}}&
                    %
				\multicolumn{1}{|c|}{P-n-P $t_\theta$} & \multicolumn{1}{c|}{0.56 $\pm$ 4.46}
                    & \multicolumn{1}{c|}{-2.75 $\pm$ 6.47}  &  \multicolumn{1}{c}{0.97 $\pm$ 7.71}  \\ \cline{2-5}
				&
				\multicolumn{1}{|c|}{P-n-P $h_\theta$} & \multicolumn{1}{c|}{\textbf{29.39} $\pm$ 16.51}
                    & \multicolumn{1}{c|}{\textbf{20.43} $\pm$ 11.13} & \multicolumn{1}{c}{-4.97 $\pm$ 3.69}  \\ \cline{2-5}
				& 
				\multicolumn{1}{|c|}{Drag}&  	\multicolumn{1}{c|}{-4.58 $\pm$ 4.89} 
                    & \multicolumn{1}{c|}{3.69 $\pm$ 8.56}  & \multicolumn{1}{c}{-1.35 $\pm$ 4.22} \\ \cline{2-5}
				& 
				\multicolumn{1}{|c|}{Dynamic} & \multicolumn{1}{c|}{9.88 $\pm$ 10.50} 
                    & \multicolumn{1}{c|}{11.36 $\pm$ 4.23} & \multicolumn{1}{c}{\textbf{11.99} $\pm$ 11.96}  \\
                    \Cline{1pt}{1-5}
			\end{tabular}
 }
	\label{tab:real-world-results}
\end{table}

Finally, we evaluate in the real-world three manipulation primitives, transferring the networks in a zero-shot manner.
The performance is reported over 3 test episodes, with 10 episode steps each, resulting in at least 30 interactions per cloth, where we discard action steps in which the grasp was unsuccessful and the cloth configuration was not altered by the grasp attempt.
Table~\ref{tab:real-world-results} shows the results of the normalised coverage improvement after 10 interactions with the cloth.
The results on the dynamic manipulation primitive show that QDP can increase the coverage improvement up to 11.99\% compared to MVP.
In addition, the P-n-P primitive where the optimal height is determined by QDP outperforms all the other manipulation primitives, for both the towel and napkin cloths, increasing the mean coverage more than 4.46\% compared to MVP P-n-P, which is the second best primitive.
These results show that modifying the velocity and height of the manipulation primitives based on the cloth state is beneficial, following the results achieved in simulation.
Additionally, the results for each primitive in the chequered rag using QDP show superior performance compared to the performance of MVP. We hypothesise that the expected transitions of the rag when performing the manipulation primitives is closer to the training data, and thus is more in line with the simulation results.  

Furthermore, Figure~\ref{fig:exp-real-world} provides qualitative results and shows the transitions for each manipulation primitive, both for QDP and MVP. During the transition from $\s_t$ to $\s_{t+1}$, the QDP dynamic manipulation primitive is executed with a value of $\v=0.3$, and QDP P-n-P uses a value of $h_\theta=0.4$. These values that are away from the median of $\v=0.1$ and $h_\theta=0.2$ enable the displacement of a larger number of cloth points, not only increasing the coverage, but also moving the cloth to a configuration more suitable for a completely flattened state.
In contrast, even though the baseline shows relative improvements in the coverage, it might get stuck in a sub-optimal cloth configuration where the cloth is too crumpled and it cannot be solved by using the fixed primitive parameter values. 

Overall there is no improvement in determining the optimal time for the drag and P-n-P primitives for all the cloths. The decreased performance of QDP for these primitives compared to the simulation results shows the sim-to-real gap for finding the optimal pick and place locations. This could be solved by retraining the network using real data instead of transferring the networks directly from simulation.
Another limitation when transferring to the real-world has been the amount of failed grasps, which could be improved by using a more accurate simulator.
We will investigate if the performance can be improved by adding as well a depth image as information in the sequential process.

\section{Conclusion}
We presented QDP, a novel approach for sequentially choosing parameter values of manipulation primitives for cloth manipulation.
While prior work has overlooked the effect of parameters such as the velocity or height of manipulation primitives, the proposed sequential decision process allows a greater variety and complexity of primitives to be used. 
This variety makes it possible  to handle diverse fabric materials and sizes.
Our experimental results show that, compared to previous work, QDP can improve up to a 20\% coverage in simulation for the task of cloth unfolding.
Furthermore, real-world experiments demonstrate the effectiveness of finding the optimal velocity and height for dynamic and quasi-static manipulation primitives.

While we demonstrated the benefit of choosing the primitive parameter values for manipulating multiple cloths, 
generalising to a broader set of materials and shapes remains an open problem.
Recent research by Hietala et al.~\cite{Hietala_2022_ours} indicates that closed-loop feedback significantly improves the adaptation capabilities in cloth manipulation.
Thus, combining parameterised primitives with real-time closed-loop feedback appears as a promising avenue towards more general and adaptive skills.

This work paves the way to a broader range of complex manipulation primitives, eliminating the human effort of fine-tuning or designing primitives, while reducing computational requirements due to the sequential decision process.
This gives promise to exploring complex parameterised manipulation skills for shaping other deformable materials such as visco-elastic or elasto-plastic ones, which are present in many industrial and house-hold environments.

\section*{Acknowledgements}
The authors would like to thank Kevin Sebastian Luck of Aalto University for their help and support in this work. The authors would also like to acknowledge the computational resources provided by the Aalto Science-IT project.

\bibliography{manipulation}
\bibliographystyle{IEEEtran}

\end{document}